\documentclass[11pt]{article}

\usepackage[final]{acl}

\usepackage{times}
\usepackage{latexsym}

\usepackage[T1]{fontenc}
\usepackage[utf8]{inputenc}

\usepackage{microtype}

\usepackage{inconsolata}

\usepackage{graphicx}

\usepackage{booktabs}
\usepackage{amssymb}
\usepackage{pifont}
\usepackage{tikz}
\usetikzlibrary{shapes.geometric, arrows.meta, positioning, calc}
\usepackage{enumitem}

\PassOptionsToPackage{svgnames}{xcolor}
\usepackage[svgnames,dvipsnames]{xcolor}

\usepackage[most]{tcolorbox}
\usepackage{fvextra} % verbatim on steroids
\DefineVerbatimEnvironment{PromptVerb}{Verbatim}{
  breaklines=true,
  breakanywhere=true,
  fontsize=\footnotesize,
  breaksymbol={},        % no symbol at line breaks
  breaksymbolleft={},    % nothing on the left side
  breaksymbolright={},   % nothing on the right side
  breaksymbolindent=0pt  % no extra indent for wrapped lines
}

\title{InSight: A Benchmark for Agentic Claim Verification in Interactive Visualizations}

\author{
  \textbf{Maeve Hutchinson\textsuperscript{1}},
  \textbf{Syed Mahbubul Huq\textsuperscript{1}},
  \textbf{Mohammad Albinhassan\textsuperscript{2}},
\\
  \textbf{Radu Jianu\textsuperscript{1}},
  \textbf{Aidan Slingsby\textsuperscript{1}},
  \textbf{Pranava Madhyastha\textsuperscript{1,3,2}}
\\
\\
  \textsuperscript{1}City, University of London,
  \textsuperscript{2}Imperial College London,
\\
  \textsuperscript{3}The Alan Turing Institute
\\
  \small{
    \textbf{Correspondence:} \href{mailto:maeve.hutchinson@city.ac.uk,pranava.madhyastha@city.ac.uk}
         {\{maeve.hutchinson, pranava.madhyastha\}@city.ac.uk}
  }
}

\begin{document}
\maketitle
\begin{abstract}
%This document is a supplement to the general instructions for *ACL authors. It contains instructions for using the \LaTeX{} style files for ACL conferences.
%The document itself conforms to its own specifications, and is therefore an example of what your manuscript should look like.
%These instructions should be used both for papers submitted for review and for final versions of accepted papers.
Vision Language Models have demonstrated remarkable proficiency in interpreting static visual artifacts, but modern data analysis is inherently dynamic, requiring the active interrogation of interactive environments. Existing benchmarks are predominantly constrained to static imagery and one-shot question answering and fail to capture the epistemic demands of this domain, where evidence is frequently occluded, distributed across linked views, or conditionally revealed through user agency. In this paper, we introduce InSight, a benchmark for agentic claim verification over interactive visualizations. The dataset consists of 21,349 claims derived from human-authored analytical narratives and grounded in fully interactive web-based environments.
%Where we reframe visual reasoning as a sequential process of evidence acquisition. 
Agents must navigate these environments to determine whether a natural language claim is supported, refuted or not verifiable given the available evidence. Unlike traditional evaluations, InSight treats interaction traces as intrinsic proxies for reasoning, enabling a rigorous audit of how models seek and synthesize visual evidence. We evaluate state-of-the-art models, revealing that interactive verification remains a non-trivial challenge. We release InSight at \url{https://github.com/maevehutch/insight}.
% Our findings demonstrate that interaction is indispensable for falsification, and that successful agents exhibit distinct, traceable strategies of visual exploration absent in their static counterparts.

\end{abstract}

\section{Introduction} %1.5 pages

The recent advances of %multimodal large language models, especially 
vision-language models (VLMs) have shown impressive performance in chart interpretation and visual reasoning \citep[e.g.,][]{google_new_2025, openai_introducing_2025}. These systems are being increasingly integrated into analytical workflows, evolving from mere curiosities into fully fledged intelligent assistants tasked with conducting visual analysis, generating design advice and captioning complex imagery \cite{kim_how_2025, kaur_chartagent_2025}. However, while the deployment of these models has indeed accelerated, the premise that they possess a robust, ground-level understanding of visual data remains contentious. Recent empirical evidence suggests that despite their fluency, VLMs frequently lack fundamental \emph{visualization literacy}, which is the ability to accurately read, understand, and interpret encoded data without relying on parametric knowledge. \citet{hong_llms_2025} demonstrate that state-of-the-art models often fail to ground their answers in visual features, instead confabulating responses based on textual context or prior training data, particularly when confronted with modified or decontextualised charts.

This discrepancy between perceived capability and actual visual grounding is exacerbated by the limitations of current evaluation methodologies. The evaluation of these models has traditionally relied on static benchmarks, where the model is presented with a fixed image of a chart and queried via a question-answer pair \cite{methani_plotqa_2020,masry_chartqa_2022} or a summarisation task or captioning task \cite{tang_vistext_2023}. While valuable for assessing perceptual acuity in closed settings, this static paradigm ignores the fundamental ecological validity of the process of data analysis where meaningful visualizations are rarely static artifacts. They are interactive environments, such as dashboards, faceted browsers, and exploratory tools, where information is conditionally revealed through user agency. In real-world analytical scenarios, evidence is frequently occluded behind tooltips, distributed across linked views, or accessible only through filtering and zooming. 
% Previous methodologies in this direction, especially in the context of NLP and the broader automated understanding of charts and chart like data, have operated solely on a static snapshot which is epistemically limited as it cannot verify what it cannot see.  

In this paper, we introduce \emph{InSight}, a benchmark for agentic claim verification over interactive visualizations. InSight reframes visual reasoning as a sequential decision-making process. We present an evaluation framework where an agent must actively navigate web-based visualization environments through executing actions such as clicking, hovering, and scrolling to verify data-related claims. Our work departs from most previous attempts in building visualization understanding benchmarks in two key ways. Firstly, unlike datasets that rely on synthetic templates or task setups \cite{methani_plotqa_2020, masry_chartqa_2022}, or crowd-sourced descriptions of static images \cite{akhtar_chartcheck_2024}, InSight is fundamentally grounded in human-authored analytical notebooks focused on real, analytical, data exploration. We extract claims directly from expert narratives and preserving the corresponding custom vega-lite visualizations \cite{satyanarayan_vega-lite_2017}, allowing us to preserve authentic analytical workflows rather than contrived proxies.
Secondly, InSight treats the \emph{interaction trace} as a first-class citizen. In traditional text-based fact-checking \cite{thorne_fever_2018, aly_feverous_2021} or static visual question answering \cite{methani_plotqa_2020, masry_chartqa_2022}, the reasoning process is often opaque, inferred only from post-hoc interpretability methods or chain of thought as a proxy. In InSight, the sequence of actions, such as, the mouse interaction or the navigation path taken, provides an explicit, traceable proxy for the model's attentional focus and reasoning strategy. This allows us to determine not only \emph{if} a model is correct, but \emph{whether} it engaged with the necessary evidence to justify its conclusion.
Our experiments with state-of-the-art models reveal that interactive verification remains a difficult challenge for all models. We also demonstrate that interaction is particularly critical for falsification, where targeted visual evidence is required to overturn linguistically plausible but unsupported claims. We also observe that high-performing models exhibit distinct interaction signatures, engaging in deeper, more targeted exploration of the visual topology.

We make the following contributions: We formulate the task of \emph{Interactive Visual Claim Verification}, extending the epistemic requirements of fact-checking to agentic, multimodal environments. We present \emph{InSight}, an ecologically valid benchmark comprising over $21k$ claims grounded in human-authored interactive visualizations, explicitly designed to penalise passive perception. We provide a comprehensive evaluation of current VLMs, offering a granular analysis of how interaction strategies correlate with reasoning fidelity.

\section{Background} % 1 page
\subsection{Agentic Vision-Language Models}
% - action space, enviroment etc., assistants (OS world etc.)
% - task-based - not about reasoning (can we use agentic VLM to trace reasoning? -- not just task completion)
% - models are general purpose, extremely capable -> becoming assistants, if we are to trust them to do tasks, need to know WHY they make certain decisions, need some reasoning trace

Recent work has explored VLMs deployed as agents that operate in interactive environments, such as web browsers, desktop GUIs, and simulated interfaces. Benchmarks such as OSWorld \cite{osworld}, MiniWoB++ \cite{furuta_multimodal_2024}, VisualWebArena \cite{koh_visualwebarena_2024} and BrowserGym \cite{chezelles_browsergym_2025} frame perception and action as a sequential decision-making problem, where models must execute actions (e.g., clicking, scrolling, typing) to complete complex tasks. These efforts have driven rapid progress in general-purpose multimodal agents and highlighted the challenges of long-horizon planning, grounding, and robustness in realistic environments.

However, existing agentic benchmarks primarily evaluate task completion, with success defined by whether a goal state is reached. In contrast, our work focuses on a single, tightly scoped task, visual claim verification, where interaction is used explicitly to acquire evidence, enabling controlled analysis of how agents seek, evaluate, and integrate information across modalities.

\subsection{Chart Understanding}
A large body of work addresses question answering and reasoning over charts and visualizations. Early datasets such as FigureQA \cite{kahou_figureqa_2018} and DVQA \cite{kafle_dvqa_2018} rely on synthetically generated charts and templated questions, enabling controlled evaluation of specific reasoning skills but offering limited ecological validity. Later datasets, including PlotQA \cite{methani_plotqa_2020} and ChartQA \cite{masry_chartqa_2022}, move toward greater realism by using real or semi-real data and more diverse visual styles, but still frame the task as one-shot question answering over static images.

Datasets such as ChartCheck \cite{akhtar_chartcheck_2024} focus on chart reasoning through claim verification rather than QA, requiring models to determine whether a textual statement is supported or contradicted by a chart. This shift aligns more closely with real-world misinformation scenarios, but these datasets continue to rely on static visual inputs and typically involve claims written or mutated specifically for the benchmark. As a result, they do not capture how analysts actually interact with visualizations to uncover evidence, nor do they reflect the compositional, multi-view, and interactive nature of real analytical dashboards.

InSight differs from prior chart datasets in two key respects: it preserves interactive visualization environments rather than static renderings, and its claims are grounded in human-authored analytical narratives, rather than being synthetically constructed to fit a predefined schema. This allows the benchmark to capture realistic tasks that are absent from existing chart understanding datasets.

\subsection{Claim Verification and Fact-Checking}
Claim verification has been extensively studied in text-only settings, most notably in datasets such as FEVER \cite{thorne_fever_2018}, SciFact \cite{wadden_fact_2020}, and FEVEROUS \cite{aly_feverous_2021}. These benchmarks emphasize evidence retrieval and textual entailment, introducing the Not Enough Information (NEI) label to capture epistemic uncertainty. In these datasets, relevant evidence is explicitly available in textual form and do not model the process of actively uncovering evidence and updating beliefs about claims.

Multimodal fact-checking datasets extend this paradigm to images. Datasets such as NewsCLIPpings \cite{luo_newsclippings_2021} and AVerImaTeC \cite{cao_averimatec_2025} require models to assess the consistency across textual and image modalities often in closed-world settings where all evidence is contained in the input. These tasks emphasize cross-modal alignment and visual-textual reasoning but still rely on static inputs and do not support sequential evidence acquisition.

InSight bridges these lines of work by combining the epistemic framing of claim verification, including NEI supervision, with interactive visual environments. In this setting, the environment is partially observable with evidence distributed across views, or conditionally revealed through interaction, introducing challenges that are absent from both text-only verification and static multimodal fact-checking.

% \subsection{visual/chart question answering? - vision explainability}
% - what traceability approaches exist? saliency maps problematic, token-level stuff problematic --- a lot of explainability approaches only work for open models, require more computation
% - intead of post-hoc explainability, why not build benchmarks with traceability in mind? -> agentic

\section{InSight: Task and Dataset} % 3 pages 
% - about dataset: what do samples look like, why
% - vis and text are collected from bla bla, ecologically valid, people with vis training
% - grounded in text alone not visual context (following changpinyo)

% language stuff here instead 
% 
% language bridging modalities

\subsection{Task Definition} \label{sec:task}
We define interactive claim verification over visualizations as a decision-making task in which a model or a human must determine the veracity of a natural language claim, $c$, in relation to an interactive visualization environment, $\mathcal{E}$. 

Each instance in the dataset consists of a natural language claim made regarding the data presented in the environment, and the environment itself. The latter is rendered as a web-based interface containing multiple views (e.g., a dashboard) and interactive UI elements.
%visualizations, and interactive visualizations, rendered in a web-based environment that may contain multiple views, like a dashboard, and interactive UI elements like filters, tool tips, and radio buttons.

The objective of the task is to assign one of three labels to the claim: %claims have one of three labels: 
\emph{True}, meaning the visualizations support the claim; \emph{False}, meaning the visualizations contradict the claim; and \emph{Not Enough Information (NEI)} meaning that the visualizations do not provide sufficient evidence to either verify or refute the claim. %The objective of the task is to output one of these labels.

%The visualization environment is not fully observable from a single static view. 

The environment $\mathcal{E}$ is partially observable. Relevant visual evidence may be occluded behind interactions, distributed across multiple views or panels, or conditionally revealed based upon prior states (of interactions). Thus, to access this evidence, the model must execute a sequence of actions $a_1, ..., a_T$ from a fixed action space (e.g., clicking, hovering, scrolling) to update the environment state and generate new visual observations. The model operates within a turn-based interaction loop with a bounded budget of actions. We emphasise that no textual metadata regarding the underlying data is provided; all verification must be grounded in pixel-level evidence revealed through interaction.
%the model may perform actions that update the environment state and produce a new visual observation. The model has a fixed action space that includes clicking, hovering and scrolling. The model operates in a turn-based interaction loop and may perform a bounded number of actions before producing a final answer. There is no additional textual evidence about the underlying data provided beyond what is visually available in the environment. All verification must be grounded in evidence revealed through interaction with the visualizations. 

We formalize this interactive claim verification task as a process of \emph{sequential evidence acquisition}. %Each action updates the state of the visualization environment, potentially exposing new information relevant to the claim. 
%In this way, the resulting action sequence reflects what information the model considers relevant to the veracity of the claim, how it navigates the visualization to obtain that information, and when it judges the available evidence to be sufficient to make a decision. In this formulation, interaction traces are intrinsic to task performance. 
As each action potentially exposes new information, the resulting interaction trace serves as an intrinsic proxy for the model's internal reasoning process, revealing what information the model deems relevant, how it navigates the topology of the visualization, and the stopping criteria it employs when judging evidence sufficiency.
% and reveal something about reasoning?? or like are intrinstic to task peformance not just procedural to complete the task? - how about that? 

% \begin{figure*}
% \begin{center}
%   \includegraphics[width=\textwidth]{figures/acl-method-fig.png}
%   \label{fig:method}
%   \caption{Worked example of the dataset construction pipeline. Starting from (A) human-authored analytical text about traffic accidents (B), we extract semantically labeled spans (C), decompose them into atomic propositions with resolved references (D1), and generate False claims through antonym substitution (4) and in-lexicon argument replacement (D2), as well as NEI claims through out-of-lexicon substitution (D3). NLI scores validate that mutations produce the intended semantic relationships.}
% \end{center}
% \end{figure*}

\subsection{Dataset Construction} \label{sec:construction}

InSight is derived from a corpus of human-authored analytical notebooks that combine interactive Vega-Lite \cite{satyanarayan_vega-lite_2017} visualizations with natural language narratives describing the analytical insights they reveal \cite{wood_design_2019}. These notebooks were authored independently by analysts with formal training in data visualization, who selected their own datasets, formulated analytical questions, and designed custom visualizations. As a result, the corpus reflects authentic analytical reasoning rather than constructed descriptions. We apply rigorous filtering criteria, retaining $N = 297$ notebooks for downstream processing.

The construction pipeline proceeds in four stages. Full implementation details, including all prompts, thresholds, and lexicons, are provided in Appendix~\ref{a:construction}.

\paragraph{Stage 1: Span Extraction.} We extract candidate spans from each narrative using \citet{lundgard_accessible_2022}'s four-level semantic model, retaining only spans at levels 2 (statistical insights) and 3 (visual/perceptual insights), which express content verifiable against visualizations. To ensure extraction stability, we perform three independent LLM runs per text segment \cite{wang_self-consistency_2023} and consolidate spans via a multi-stage agreement procedure based on trigram overlap. Spans extracted by only a single run are discarded, and the final span text is constructed from tokens selected by at least two runs. This procedure filters out hallucinated extractions and semantically irrelevant text.

\paragraph{Stage 2: Claim Decomposition.} Retained spans may contain compound statements or implicit references. Following \citet{metropolitansky_towards_2025}, we decompose each span into atomic, fully decontextualized propositions that can be independently verified against the visualizations. A self-verification step confirms that each decomposed claim is entailed by the source text \cite{weng_large_2023}, and semantic labels are reassigned via majority vote across three passes. Claims that fail grounding verification or exhibit unstable labeling are discarded. The resulting claims constitute the ground-truth \textsc{True} set.

\paragraph{Stage 3: Claim Mutation.} To construct a balanced verification benchmark, we generate \textsc{False} and \textsc{NEI} claims through controlled mutation of the true claims, following the contrastive construction paradigm of \citet{thorne_fever_2018}. We apply three mutation strategies, each targeting a distinct source of semantic change:

\begin{itemize}[nosep]
    \item \textbf{Antonym substitution} reverses directional or quantitative language (e.g., \emph{increased} $\to$ \emph{decreased}) using a curated antonym lexicon, producing \textsc{False} claims.
    \item \textbf{In-lexicon argument substitution} replaces data arguments (categorical values, numerical quantities) with alternatives from the same dataset, producing \textsc{False} claims.
    \item \textbf{Out-of-lexicon argument substitution} replaces arguments with values or attributes absent from the dataset, producing \textsc{NEI} claims that refer to entities for which no visual evidence exists.
\end{itemize}

\noindent All mutations are constrained by XML-style tagging of substitutable spans to prevent lexical drift and preserve structural similarity with the source claim. Crucially, each mutated claim is validated using a pretrained NLI model \cite{he_debertav3_2023}: \textsc{False} claims must exhibit high bidirectional contradiction scores ($> 0.9$) with the original, while \textsc{NEI} claims must exhibit low entailment \emph{and} low contradiction ($< 0.2$). Claims failing validation are discarded.

\paragraph{Stage 4: Human Validation.} We conduct a human validation study with 13 expert annotators, who produced 475 annotations across 294 sampled claims. Annotators performed the same interactive verification task defined in \S\ref{sec:task}, blind to the original labels and construction process. Raw agreement with the dataset labels is 81.3\%. A Bayesian Dawid--Skene model \cite{dawid_maximum_1979, paun_comparing_2018} applied to the annotation data recovers class prevalences within 3 percentage points of the ground truth, with a posterior mean accuracy of 66.2\% (95\% CI: [59.5\%, 71.8\%]) and an MAP estimate of 77.6\%. These results confirm that the controlled construction and NLI-based validation procedures produce labels broadly consistent with expert human judgment, despite the inherent difficulty of the task.

\subsection{Dataset Analysis} \label{sec:analysis}
 
InSight contains 21,349 claims derived from 297 human-authored analytical notebooks. Claims are distributed across the three verification labels, with 41.6\% True, 45.0\% False, and 13.4\% Not Enough Information (NEI) cases.
 
\noindent\textbf{Claim properties.} Claims are concise but non-trivial, with an average length of 18 tokens (median 17). 56.8\% correspond to semantic level 2, relating to statistical and numerical insights (e.g., value retrieval, extrema), and 43.2\% to level 3, relating to visual and perceptual insights (e.g., trends, patterns). This distribution reflects a mix of precise data lookup and higher-order relational reasoning, both of which are common in real analytical workflows.
 
\noindent\textbf{Visualization diversity.} The visualizations are authored in Vega-Lite \cite{satyanarayan_vega-lite_2017}, a declarative grammar based on composable marks and encodings. As a result, the space of visualization forms is not bounded by a predefined taxonomy. Across the corpus, visualizations combine a wide range of marks---bars, lines, points, areas, geoshapes, text, and rules---and frequently involve layering and multi-view composition. On average, each notebook contains 4.5 views and 3.7 visualization specifications, with 67\% of notebooks exhibiting explicit multi-view composition. This means that part of the verification task is determining \emph{which} visualization is most relevant to the claim, a challenge absent from single-chart benchmarks.
 
\noindent\textbf{Interaction characteristics.} Interactivity is intrinsic to the visualization environments in InSight. The most prevalent mechanism is hover-triggered tooltips, which conditionally reveal details-on-demand, such as precise values, labels, or additional information. Click-based and point-selection interactions are also common, enabling filtering, highlighting, or cross-view coordination. Less frequent primitives (interval selection, zoom/pan) nevertheless play an important role for claims requiring range-based or spatial reasoning. Importantly, these statistics capture interaction capabilities designed into the visualizations themselves; scrolling and basic mouse actions are additionally available as part of the agent action space (\S\ref{sec:action_space}).

\noindent\textbf{Partial observability and interaction necessity.} Every environment in InSight is partially observable, and therefore every claim requires exploration of the environment. Environments contain on average 4.5 views and 3.7 specifications (Figure~\ref{fig:corpus_stats}), and cannot be rendered in a single fixed viewport, so the initial screenshot never exposes the full set of visualizations. Even when the visualization relevant to a claim happens to appear first, an agent cannot know this without inspecting the remaining views: it must establish that no other view supplies stronger, contradicting, or disambiguating evidence, and for NEI claims it must establish that no view supplies evidence at all. In addition, the most prevalent interaction mechanism, hover-triggered tooltips, hides precise values behind an action, so even claims about a visible chart typically cannot be resolved from the static rendering. 

Additionally, visualizations here were built by analysts to \emph{explore} their own datasets, and the claims were derived from insights they reached through interactive use, not from what a single frame conveys. Consequently, no evaluation claim is verifiable from the initial viewport alone, and a correct answer produced without interaction should be attributed to parametric knowledge or guessing rather than to visual grounding. This property is what allows InSight to penalize passive perception, and it motivates the design of the Interaction Efficiency Score in \ref{sec:metrics}.
 
\noindent\textbf{Comparison with prior datasets.} Unlike existing chart understanding and claim verification benchmarks, which typically rely on static images or pre-extracted data, InSight jointly supports: (i) fully interactive visual environments, (ii) claims grounded in real-world analytical provenance rather than synthetic construction, (iii) explicit NEI supervision, and (iv) the collection of action traces during verification. See Appendix~\ref{a:dataset} for additional distributional analyses, including mark-type frequencies and specification density histograms.

\section{Evaluation and Behavioural Analysis} % up to a page

% We evaluate VLMs on the InSight benchmark by deploying them as interactive agents that must verify claims as defined in Section~\ref{sec:task}.

\begin{figure*}[t]
\begin{center}
  \includegraphics[width=\textwidth]{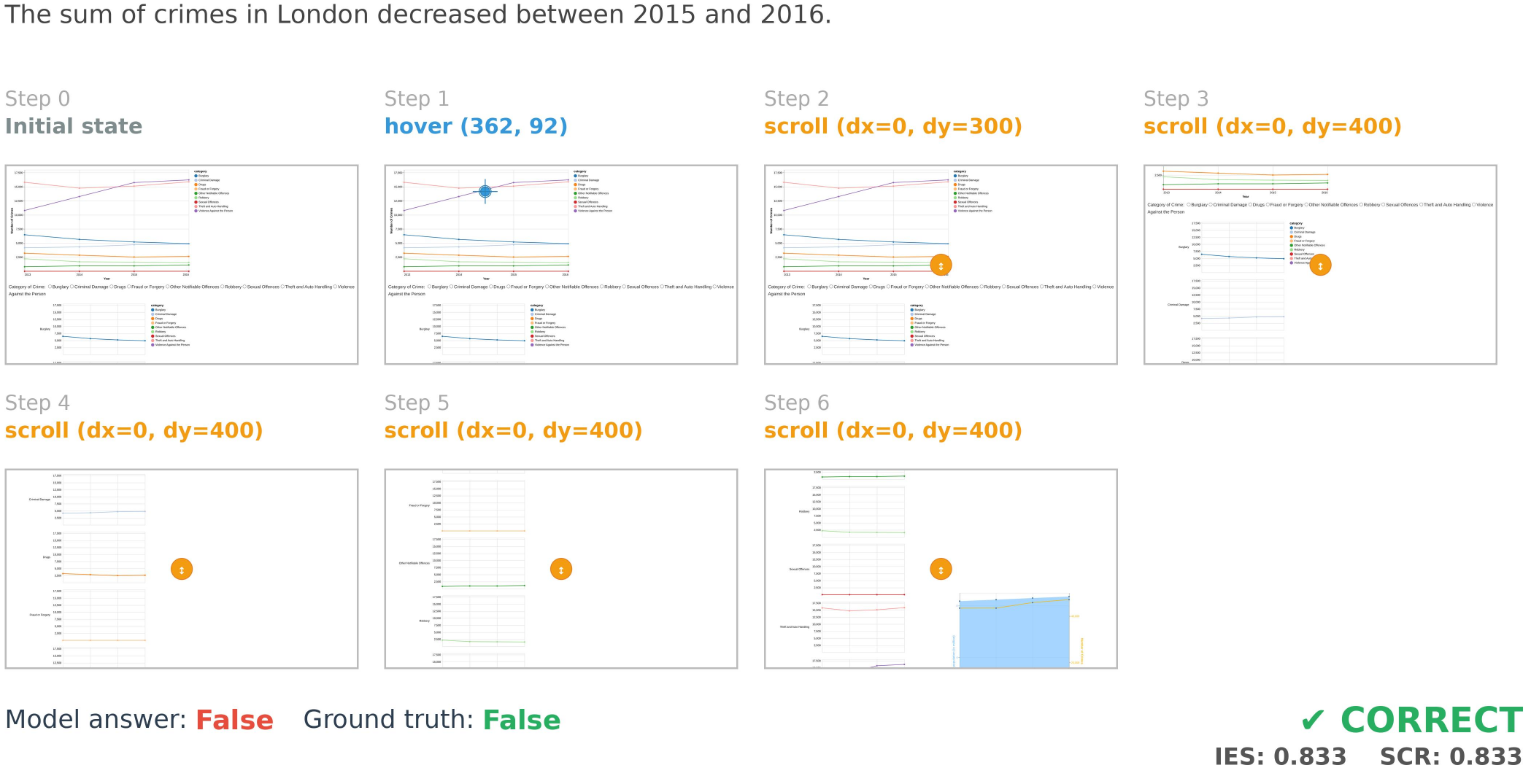}
  \caption{Interaction trace for Gemini 3.5 Flash verifying a false claim about London crime trends. The model hovers to inspect a tooltip, then scrolls through the multi-view notebook to locate relevant evidence, correctly predicting \textsc{False} with an IES of 0.833.}
  \label{fig:trace_example}
\end{center}
\end{figure*}

\subsection{Experimental Setup} \label{sec:action_space}

\noindent\textbf{Evaluation Environment.} Each visualization environment $\mathcal{E}$ is rendered as an HTML page containing one or more interactive visualizations authored in Vega-Lite and executed in a Chromium-based headless browser using Playwright with a fixed viewport. This setup renders the visualization notebooks exactly as a human expert would see them and ensures consistent rendering and interaction behavior across models. The fixed viewport is deliberately smaller than the rendered page: because notebooks contain multiple views (\S\ref{sec:analysis}), only a portion of the environment is visible at any time, and the remainder must be reached by scrolling, while details-on-demand (tooltips, selections) must be triggered by mouse actions. Every environment is therefore partially observable, and no claim in the evaluation set can be verified from the initial screenshot alone, regardless of which visualization contains the relevant evidence.

Crucially, we treat the web page as a purely visual surface. The model is provided only with the RGB screenshot of the current viewport; it does not have access to the underlying DOM tree, HTML source code, or accessibility tree. This design choice aligns with recent work in GUI agents \citep[e.g.][]{uitars, luo_visual_2025}, ensuring that the model must ground its reasoning in the visual rendering of data rather than parsing textual data representations embedded in the code. 

% \subsection{Agent Action Space}

\noindent\textbf{Agent Action Space.} The model interacts with the environment through a defined high-level action space $\mathcal{A}$, drawing on prior work in web and UI agent benchmarks \citep[][interalia]{osworld, uitars}. The action space mimics standard mouse and keyboard inputs, requiring the model to predict pixel coordinates $(x, y)$.

We define the action space $\mathcal{A}$ as follows:
\noindent\textit{Navigation:} \texttt{scroll(dx,dy)} for precise pixel scrolling, and macro-actions such as \texttt{page\_up(n)}, \texttt{page\_down(n)}, and directional arrow keys (e.g., \texttt{arrow\_right(n)}) to navigate the webpage.
\noindent\textit{Mouse Interaction:} \texttt{drag((x1,y1), (x2,y2))}, \texttt{click(x,y)}, \texttt{shift\_click(x,y)}, and \texttt{hover(x,y)}. These are essential for interacting with visualizations to reveal more evidence, such as tooltip inspection (hover), range selection (drag), or multi-selection (shift-click).
\noindent\textit{Termination:} \texttt{answer(label)}, where label $\in \{\text{True}, \text{False}, \text{NEI}\}$. This action terminates the episode and submits the verification decision.

% We acknowledge that while this action space mirrors prior work, it differs from naturalistic human behavior in certain respects. For instance, humans perform continuous mouse movements and may hover implicitly while navigating, whereas our action space discretizes these into individual parameterized actions. However, this discretization is necessary to keep interaction traces tractable and interpretable. An alternative approach involving continuous low-level actions would yield intractably long sequences that would be difficult to analyze or reproduce.

% \subsection{Interaction Protocol}

\noindent\textbf{Interaction Protocol.} Models interact with the environment through a standard turn-based protocol. At each turn $t \leq T_{\text{max}}$, the model receives the current screenshot $s_t$ and the claim $c$, and must output a single action $a_t \in \mathcal{A}$. If $a_t$ is not a terminal \texttt{answer} action, the environment executes $a_t$, updates its state, and provides the next screenshot $s_{t+1}$. The process continues until the model outputs an \texttt{answer} action or the maximum turn budget $T_{\text{max}}$ is reached. 

% Each episode is bounded to $T_{\text{max}} = 10$ across all experiments, similar to prior work \citep[e.g.][]{osworld, wu-etal-2025-webwalker}.

\subsection{Evaluation Metrics} \label{sec:metrics}
We introduce an \textit{Interaction Efficiency Score} (IES), designed to capture whether agents interact productively with the visualization environment while performing claim verification. Existing agentic benchmarks primarily evaluate task completion alone, whereas InSight explicitly treats interaction traces as part of the reasoning process. Our metric is conceptually related to Success weighted by Path Length (SPL) \cite{spl} used in embodied navigation benchmarks, but adapted to browser-based visual evidence acquisition.

For an episode $i$, let $A_i \in [0,1]$ denote answer accuracy, where $A_i = 1$ if the predicted label is correct and $A_i = 0$ otherwise. Let $T_i$ denote the total number of actions executed by the agent, and let $E_i$ denote the number of effective actions. We define an effective action as an action that produces an observable change in the environment state, i.e., interactions that reveal new visual evidence.

Formally, we compute effective actions as:

\begin{equation}
E_i = \sum_{t=1}^{T_i} \mathbf{1}[s_{t+1} \neq s_t]
\end{equation}

where $s_t$ denotes the observable environment state after action $t$, and $\mathbf{1}[\cdot]$ is the indicator function.

We then define the Interaction Efficiency Score as:

\begin{equation}
\mathrm{IES}_i =
A_i \cdot \frac{E_i}{T_i}
\end{equation}

where $\frac{E_i}{T_i}$ is the state change ratio (SCR) and we report the dataset-level score as:

\begin{equation}
\mathrm{IES} =
\frac{1}{N}
\sum_{i=1}^{N}
A_i \frac{E_i}{T_i}
\end{equation}

This formulation rewards agents that both answer correctly and interact productively with the environment. Incorrect answers receive a score of $0$ regardless of interaction behavior, while correct predictions are additionally differentiated by the proportion of actions that meaningfully advance the observable state of the environment. IES thus provides a behavioural signal indicating whether an agent's interaction trajectory acquires visual evidence during verification, complementing accuracy rather than replacing it.

\emph{Correct answers without interaction.} An agent that submits an answer at the first turn has $T_i = 0$ and $E_i = 0$; we define $E_i/T_i = 0$ in this case, so the episode receives IES $= 0$ even if the answer is correct. This is deliberate: since no claim is resolvable from the initial screenshot, a correct answer with no interaction reflects parametric knowledge or chance rather than visual grounding, which is precisely the failure mode IES is meant to expose. Qwen 3.5 0.8B, which never interacts, receives IES $= 0$ for this reason.

\emph{Scrolling.} Scrolling almost always changes the rendered pixels and is therefore almost always counted as effective. This is by design, because environments contain more views than fit in a viewport, scrolling is the action by which an agent discovers where relevant evidence resides and by which it returns to a relevant view after surveying the alternatives.

\emph{NEI claims.} For an NEI claim, the ideal trajectory is to explore the environment and find nothing. Such exploration still changes the observable state and is counted as effective. Since for NEI claims there is by construction no evidence directly relevant to the claim's veracity, no metric can score whether the answer \emph{depended on} revealed evidence, therefore IES scores whether the agent looked. The same holds to a lesser degree for True and False claims, because the agent cannot know in advance which view is relevant, actions that reveal irrelevant views are still useful exploratory work in a partially observable environment, and we intentionally do not penalize them.

\emph{Limitation.} IES credits any state-changing action regardless of whether the final answer used the revealed evidence. It therefore cannot fully rule out correct-but-ungrounded predictions by agents that interact but ignore what they see. IES identifies agents that do not interact at all, and ranks interacting agents by how much of their budget produces new observations. Finer-grained grounding would require claim-specific evidence annotations and is a natural extension we leave for future work.

\begin{table*}[t]
\centering
\caption{Comparison of model efficiency and interaction metrics across model families. Accuracy and IES are reported overall and disaggregated by claim label.}
\label{tab:model_comparison}
\resizebox{\textwidth}{!}{%
\begin{tabular}{lc|c|c|c c c|c|c|c c c}
\toprule
\textbf{Model} & \textbf{$T_{max}$} & \textbf{$T/N$} & \textbf{Accuracy} & \textbf{Acc$_{\text{False}}$} & \textbf{Acc$_{\text{NEI}}$} & \textbf{Acc$_{\text{True}}$} & \textbf{$E/T$} & \textbf{IES} & \textbf{IES$_{\text{False}}$} & \textbf{IES$_{\text{NEI}}$} & \textbf{IES$_{\text{True}}$} \\
\midrule
Gemini 3.5 Flash       & 25 & 9.33  & 50.0\% & 54.8\% & 50.6\% & 44.6\% & 56.7\% & 31.0\% & 29.87\% & 36.10\% & 27.03\% \\
Gemini 3.5 Flash       & 10 & 6.27  & 41.6\% & 45.8\% & 45.8\% & 33.3\% & 61.3\% & 24.93\% & 23.62\% & 32.63\% & 18.63\% \\
Gemini 3.5 Flash     & 1 & - & 44.2\% & 33.1\% & 73.5\% & 26.2\% & - & - & - & - & -\\
GPT-5.5                & 10 & 3.13  & 57.2\% & 59.6\% & 59.0\% & 53.0\% & 45.4\% & 26.98\% & 29.24\% & 23.98\% & 27.70\% \\
\midrule

Gemma 4 31B            & 10 & 3.16  & 47.2\% & 40.4\% & 63.9\% & 37.5\% & 43.8\% & 20.57\% & 15.12\% & 28.48\% & 18.13\% \\
Gemma 4 31B          & 1 & -  & 45.4\% & 26.5\% & 84.9\% & 25.0\% & - & - & - & - & - \\
Gemma 4 E4B            & 10 & 0.89  & 43.2\% & 18.7\% & 88.0\% & 23.2\% & 18.3\% & 6.87\%  & 2.81\%  & 15.66\% & 2.18\% \\
Gemma 4 E2B            & 10 & 0.16  & 31.2\% & 7.8\% & 71.7\% & 14.3\% & 5.0\% & 0.90\%  & 0.00\%  & 2.71\%  & 0.00\% \\
\midrule

Qwen 3.6 35B           & 10 & 1.78  & 44.8\% & 42.8\% & 62.7\% & 29.2\% & 21.9\% & 9.07\%  & 7.54\%  & 12.67\% & 7.01\% \\
Qwen 3.5 27B           & 10 & 1.57  & 45.0\% & 38.0\% & 66.3\% & 31.0\% & 17.7\% & 7.57\%  & 5.65\%  & 10.20\% & 6.87\% \\
Qwen 3.5 27B           & 1 & -  & 46.2\% & 35.5\% & 79.5\% & 23.8\% & - & -  & -  & - & - \\
Qwen 3.5 9B            & 10 & 0.50  & 38.0\% & 50.6\% & 26.5\% & 36.9\% & 9.56\% & 3.52\%  & 4.67\%  & 3.82\%  & 2.08\% \\
Qwen 3.5 2B            & 10 & 0.24  & 37.0\% & 30.7\% & 51.8\% & 28.6\% & 4.10\% & 1.17\%  & 1.20\%  & 0.24\%  & 2.06\% \\
Qwen 3.5 0.8B          & 10 & 0.00  & 34.0\% & 21.7\% & 54.8\% & 25.6\% & 0.00\% & 0.00\%  & 0.00\%  & 0.00\%  & 0.00\% \\
\bottomrule
\end{tabular}%
}
\end{table*}

\subsection{Model Results} \label{sec:results} % 2 pages

We benchmark a selection of closed-source and open-weight state-of-the-art models. We evaluate models on a stratified subset of 500 claims, a similar sample size to equivalent multi-turn agentic benchmarking \cite{osworld}. The subset consists of 167 True, 167 False, and 166 NEI claims. We do not benchmark across the whole dataset to avoid potential data contamination issues and reserve samples for future rigorous benchmarking \cite{deng_unveiling_2024}.

We evaluate fourteen model configurations spanning four model families: Gemini, GPT, Gemma, and Qwen. For all models we use an interaction budget of 10, following other similar CUA benchmark action budgets \cite{osworld}. We additionally do ablations with interaction budgets of ($T_{\max} \in \{1, 25\}$) for some models to compare behavior across action budgets. Table \ref{tab:model_comparison} reports accuracy, interaction volume ($T/N$, the mean number of actions per episode), the effective action ratio ($E/T$), and the Interaction Efficiency Score (IES). Both accuracy and IES are reported overall and disaggregated by claim label.

\noindent\textbf{Overall Performance} At $T_{max} = 10$, GPT-5.5 achieves the highest overall accuracy at 57.2\%, followed by Gemma 4 31B (47.2\%) and Qwen 3.5 27B (45.0\%). These results are notable given that random performance on a balanced three-class task is 33.3\%, indicating that all models meaningfully exceed chance, but none approach human-level reliability. A clear scaling trend is visible within model families: Qwen accuracy degrades monotonically from 44.8\% at 35B parameters to 34.0\% at 0.8B, and Gemma drops from 47.2\% (31B) to 31.2\% (E2B). At the smallest scales, models barely exceed random baselines, suggesting that interactive visual reasoning imposes a minimum capacity threshold that sub-billion parameter models consistently fail to meet.

\noindent\textbf{The Role of Interaction} The $T_{\max}=1$ setting serves as a \emph{no-interaction baseline}. In this ablation, the model receives the claim and the initial viewport and must answer at the first turn, which reduces the task to static visual question answering over a partial view of the environment. Performance in this setting reflects what the model can infer from parametric knowledge, linguistic plausibility of the claim, and whatever happens to be visible in the first screenshot. The results are mixed. Gemini 3.5 Flash is slightly more accurate without interaction (44.2\% vs.\ 41.6\%), as is Qwen 3.5 27B (46.2\% vs.\ 45.0\%), whereas Gemma 4 31B drops from 47.2\% to 45.4\% when interaction is removed. For the two models that do not benefit, the gains at $T_{\max}=1$ are within a few points and suggest that unrestricted interaction can be counterproductive when a model lacks a plan for exploration and may commit on partial evidence, or lose track of the claim across turns. That the pattern does not hold for Gemma indicates the effect is model-specific rather than a property of the task, and it is worth noting that the accuracy gap between the two settings is small for all three models, so parametric knowledge and the initial viewport account for a large share of the accuracy that interaction-capable models achieve.

The Gemini budget ablation is informative here because it is non-monotonic: 44.2\% at $T_{\max}=1$, 41.6\% at $10$, and 50.0\% at $25$. Under the 10-action budget Gemini is the most active model, and many of its trajectories are still mid-exploration when the budget forces an answer or when the model chooses to stop. A partially completed survey of a multi-view environment appears to be worse than none, presumably because the model has seen enough to abandon its prior but not enough to replace it. Given 25 actions it uses 9.33 on average, completes more of its exploration, and reaches the highest accuracy of any Gemini configuration, with IES rising from 24.93\% to 31.0\% even though its effective action ratio falls slightly, consistent with later actions being used to return to and inspect views already discovered.

The interaction traces reveal a pronounced disparity in how models allocate their action budgets. Gemini 3.5 Flash is the most active agent, averaging 6.27 actions per episode with an effective action ratio of 61.3\%, indicating that a majority of its interactions produce observable state changes. GPT-5.5 is comparatively economical, averaging 3.13 actions but achieving the highest accuracy, suggesting more deliberate and targeted exploration. At the other extreme, Qwen 3.5 0.8B averages zero actions per episode, defaulting to immediate answer submission without any environmental interaction, a behaviour that reduces the task to uninformed guessing from the initial viewport.

\noindent\textbf{Interaction Efficiency} The IES metric disentangles accuracy from interaction quality, rewarding models that both answer correctly and interact productively. GPT-5.5 leads with an IES of 26.98\%, followed closely by Gemini 3.5 Flash (24.93\%) and Gemma 4 31B (20.57\%). The gap between accuracy and IES rankings is informative: while Gemma 4 31B achieves comparable accuracy to Gemini 3.5 Flash, its lower effective action ratio reduces its IES, indicating that a substantial proportion of its actions fail to advance the observable environment state, meaning it is not making use of the visual information.

Small models exhibit near-zero IES scores. Gemma 4 E2B achieves 0.90\% and Qwen 3.5 0.8B achieves 0.00\%, indicating that even when these models occasionally produce correct answers, they do so without meaningful environmental engagement. This finding validates the design of IES as a metric that penalizes correct-but-ungrounded predictions, which accuracy alone cannot detect.

\noindent\textbf{Label-Specific Analysis} Disaggregating IES by claim label reveals asymmetric difficulty across verification categories. For GPT-5.5, IES is relatively balanced across labels, suggesting robust interaction strategies across all claim types. In contrast, most other models exhibit a characteristic pattern where NEI claims yield the highest IES. Gemini 3.5 Flash achieves IES$_{\text{NEI}}$ = 32.63\% but only IES$_{\text{True}}$ = 18.63\%, and this gap is even more pronounced in smaller models such as Gemma 4 E4B (IES$_{\text{NEI}}$ = 15.66\%, IES$_{\text{True}}$ = 2.18\%).

This pattern admits a plausible explanation: verifying a True claim requires locating and correctly interpreting specific visual evidence, whereas recognizing that evidence is insufficient for an NEI claim may require less precise interaction; the model needs only to fail to find confirming or contradicting evidence across the environment. Conversely, False claims, which require discovering evidence that contradicts a linguistically plausible statement, are particularly challenging for weaker models, indicating difficulty to productively acquire disconfirming evidence. This is consistent with prior observations in textual fact-checking that falsification is cognitively harder than verification, and extends that finding to the agentic multimodal setting \cite{thorne_fever_2018}.

\noindent\textbf{Actions and Behavioral Signatures} Analyzing the actions taken by the models reveals qualitative differences in exploration strategies. Gemini 3.5 Flash employs the broadest action vocabulary, with substantial use of scroll (2.51 per sample), hover (1.42), click (1.15), and page\_down (0.73), as well as non-trivial use of arrow keys and drag actions. This diverse repertoire enables it to achieve the highest effective action ratio.  GPT-5.5 similarly prioritises hover (1.07) and scroll (1.02) but uses fewer total actions, consistent with a more efficient exploration strategy.

In contrast, smaller models exhibit a smaller range of action types taken. Gemma 4 E2B and Qwen 3.5 0.8B use almost exclusively zero actions across all types, while intermediate models such as Qwen 3.5 9B show limited use of scroll and click but negligible use of hover, which is a critical action for tooltip inspection. Given that hover-triggered tooltips are the most prevalent interaction mechanism in the InSight environments (Appendix \ref{a:dataset}), this omission likely explains these models' inability to access conditionally revealed evidence.

\noindent\textbf{Failure Modes} Several systematic failure patterns emerge from the interaction traces. First, \textit{premature commitment}: smaller models frequently terminate after one or two actions, submitting answers before engaging with the environment. Second, \textit{unfocused exploration}: some incorrect predictions by larger models involve many actions with low effective action ratios, suggesting exploration that fails to target relevant visual elements. Third, \textit{action-type rigidity}: mid-range models tend to default to scrolling and clicking without employing hover or drag actions, missing evidence encoded in tooltips or selection-based interactions.

These findings collectively demonstrate that interactive claim verification is not simply a harder version of static visual question answering. It requires coordinated perception, hypothesis formation, and action selection, which are capabilities that scale with model capacity but remain far from saturated even at the frontier.

% \begin{figure}
% \begin{center}
%   \includegraphics[width=1.1\columnwidth]{figures/action_type_heatmap.pdf}
%   \label{fig:actions}
%   \caption{Action type heatmap}
% \end{center}
% \end{figure}

\section{Conclusion} % 0.5 page 
% - utility of dataset more generally 
% - ecological validity 
% - agentic stuff 
% - data visualization and claim verification
% - misleading visualizations
% - visualization literacy (is limited in definition - there are different tasks, this another - in-context literacy, closer to real-life scenario)
% - multimodal argumentation
% - multimodal reasoning/planning 

% - stuff about models/action spaces 
% - 
We have presented InSight, a benchmark for agentic claim verification over interactive visualizations comprising 21,349 claims grounded in human-authored analytical notebooks. By requiring models to actively navigate web-based environments to gather evidence, InSight moves beyond static visual question answering and captures the sequential, exploratory nature of real analytical workflows.
 
Our evaluation reveals several findings. First, interactive claim verification remains challenging: the best-performing model, GPT-5.5, achieves only 57.2\% accuracy. Second, IES exposes models that answer correctly without engaging with the environment at all, which indicates ungrounded prediction rather than evidence-based reasoning. Third, falsification is disproportionately difficult, requiring targeted evidence-seeking interaction that most models struggle to produce, extending prior observations from textual fact-checking to the agentic multimodal setting.

These findings point to several directions for future work. The gap between static and interactive performance suggests that explicit training for evidence-seeking behavior, rather than one-shot visual reasoning, may be necessary. The difficulty of falsification motivates investigation into how models form and update hypotheses during multi-turn visual exploration. More broadly, InSight provides a foundation for studying multimodal argumentation in agentic systems, where the reasoning process is not opaque but traceable through interaction.

\section{Limitations}

InSight has several limitations that point to directions for future work. 
The benchmark assumes a fixed, high-level action space that, while grounded in prior work on agentic benchmarks, abstracts common mouse and navigation interactions. While this enables controlled comparison across models, it may not capture all interaction modalities used by humans, such as complex gesture-based interactions or semantic shortcuts provided by specific interfaces. Different action space designs may yield different interaction strategies and performance profiles.

Although InSight emphasizes interaction as a proxy for reasoning, interaction traces do not fully capture internal model deliberation. A model may perform correct actions for the wrong reasons or fail despite internally plausible reasoning. Thus, interaction traces should be viewed as complementary, not exhaustive, evidence of reasoning behavior. However, we conjecture that these traces still provide evidence for examining model behavior.

\section{Ethical Considerations}
%This study and its data collection procedures were formally approved by our university’s Research Ethics Committee. The origins of the dataset is a part of the student coursework

The construction of this corpus was subject to strict institutional governance and was formally ratified by the university’s Research Ethics Committee (REC). The corpus comprises artefacts originating from the capstone projects of a graduate-level module in Advanced Data Visualisation. Unlike uncurated web-scraped datasets, which often suffer from variable quality and noisy alignment, our data collection protocol imposed a rigorous high-pass quality filter. 
The inclusion criteria were determined by a two-stage validation process. Each submission was assessed by a senior academic specialising in data visualisation and visual analytics. This evaluation served as our primary signal of quality; only submissions demonstrating substantial technical and design proficiency were selected for inclusion. The initial assessment was subsequently audited and validated by the departmental Academic Board, ensuring that the selection serves as a gold-standard representation of domain concepts.
Following ethical approval, we engaged in a retrospective consent acquisition process. We contacted graduates of the programme to articulate the study’s objectives, specifically the contribution of their work to the advancement of multimodal machine learning.
The annotation procedure was also subject to ratification by the REC. We recruited data science master's students via email to participate in the study. Participation was voluntary and participants were not paid.

\section*{Acknowledgments} This work
was supported in part by the Alan Turing Institute under Fundamental Research Project No. PP00029. 

% Upon receiving approval, we contacted graduates of the program to inform them about the study’s aims and potential contributions. We obtained explicit informed consent from those who agreed to participate, specifically for the use of their coursework in our research. The dataset exclusively comprises submissions from students who voluntarily provided permission for their materials to be processed and released as part of this research.

% Bibliography entries for the entire Anthology, followed by custom entries
%\bibliography{custom,anthology-overleaf-1,anthology-overleaf-2}

% Custom bibliography entries only
\bibliography{custom, visworld}

\twocolumn
\appendix

\section{Dataset Construction Details} \label{a:construction}
 
This appendix provides full implementation details for the four-stage dataset construction pipeline summarized in \S\ref{sec:construction}.
 
\subsection{Data Sources and Filtering}
 
InSight is derived from human-authored analytical notebooks that combine interactive Vega-Lite \cite{satyanarayan_vega-lite_2017} visualizations with rich natural language narratives \cite{wood_design_2019}. These notebooks were authored independently by analysts with formal training in data visualization. They selected their own datasets, formulated analytical questions, and designed custom visualizations to explore them. The resulting notebooks reflect authentic analytical reasoning about data rather than constructed descriptions.
 
We apply a series of filtering criteria to ensure quality and relevance, retaining $N = 297$ notebooks. We extract two components from each: the human-authored analytical narrative, which serves as the source of grounded claims, and the corresponding interactive visualization environment.
 
\subsection{Span Extraction and Semantic Labeling}
 
The analytical narratives contain a mixture of descriptive context, commentary, and claims grounded in visual evidence. We isolate spans that express verifiable claims about the underlying data using an agreement-based extraction procedure.
 
\paragraph{Candidate Span Extraction.} We adopt \citet{lundgard_accessible_2022}'s four-level semantic model of natural language about data visualizations, specifically levels 2 and 3. Level 2 captures statistical insights about the underlying datasets, such as extrema or value retrieval, while level 3 captures visual insights revealed by the visualizations, such as trends or patterns. This ensures that candidate spans relate to content that could be verified or refuted by visual evidence.
 
We prompt Gemini 2.5 Flash \citep{noauthor_gemini_2025} to extract text spans from each analytical narrative containing level 2/3 semantic content. To reduce the potential impact of variability in model outputs, we perform three independent runs per text segment \cite{wang_self-consistency_2023}.
 
\paragraph{Agreement-Based Span Consolidation.} We consolidate spans across the three model runs using a multi-stage agreement procedure. Any spans that are not extractive (i.e., not verbatim substrings of the original text) are discarded. This ensures that no new content is introduced by the LLM at this stage.
 
We group spans across runs based on lexical overlap by computing the word trigram F1 overlap score and greedily matching spans from different model runs with high lexical similarity, using a fixed minimum threshold for matching. Spans that are not grouped with any other span, meaning only one model run extracted that span, are discarded.
 
From the groups of matched spans, we construct a final span by retaining tokens chosen by at least two model runs. If this produces a non-contiguous sequence, we expand it to the minimal contiguous substring in the original text. The semantic label assigned to this span is the majority label across the runs. This procedure enforces cross-run stability, filtering out hallucinated extractions and semantically irrelevant text.
 
\subsection{Span Decomposition into Verifiable Claims}
 
The spans extracted in the previous stage may contain multiple propositions, implicit references, or compound statements that are not directly suitable for claim verification. We decompose each retained span into a set of atomic, verifiable claims that can be independently assessed against the visualizations. This stage is adapted from Claimify \cite{metropolitansky_towards_2025}, tailored to interactive visual evidence.
 
We prompt an LLM to identify all specific and verifiable propositions expressed in each span. The model is instructed to decompose compound statements into their simplest discrete units and to discard non-verifiable content. Each claim must be fully decontextualized -- interpretable in isolation, without access to the surrounding narrative -- while preserving its original meaning. The model is provided with the span's immediate textual context and required to explicitly resolve all referents (e.g., pronouns, comparatives, implicit subjects). Claims are restricted to information present in the original span or its local context; no external knowledge is permitted.
 
To mitigate potential meaning shifts introduced by LLM-based decomposition, we apply a self-verification step in which the model is shown the original span and the decomposed claims and asked to verify that each claim is entailed by the source text \cite{weng_large_2023}. We also reassign semantic labels to each claim using the same semantic schema, performing 3 passes and retaining the majority label. Claims for which semantic labeling is unstable or that fail grounding verification are discarded.
 
The claims produced by this stage constitute the set of ground-truth \textsc{True} claims in InSight. Each is directly derived from human-authored analytical text, verified to preserve its original meaning after decomposition and decontextualization. These true claims form the semantic anchor of the dataset and are subsequently used as the source for controlled generation of \textsc{False} and \textsc{NEI} variants.
 
\subsection{Claim Mutation}
 
We generate \textsc{False} and \textsc{NEI} claims through controlled mutation of the true claims, applying minimal, semantics-preserving perturbations designed to alter veracity while retaining linguistic plausibility and structural similarity. This reduces annotation artifacts by ensuring that true, false, and NEI claims differ primarily in their semantic relation to the underlying visual evidence.
 
\paragraph{Antonym Substitution.} This strategy targets claims whose veracity depends on scalar, directional, or quantitative language. We construct a lexicon of lemmas that have a clear antonym which, when substituted, induces a semantically opposed claim. The lexicon is curated from high-frequency lemmas in the true claim set, focusing on auxiliary verbs, determiners, adjectives, adverbs, and nouns expressing quantitative judgments (e.g., increase/decrease, all/none). Mappings are curated to avoid substitutions that merely weaken a claim or introduce ambiguity, ensuring clear semantic opposition.
 
We tag occurrences of lexicon entries in each claim using spaCy \cite{honnibal_spacy_2020} and custom XML-style markers that encode the lemma identity. This constrains the mutation process, preventing lexical drift. We then prompt an LLM to generate a mutated false claim by substituting only the tagged tokens with an appropriate antonym, preserving all other content verbatim and making only minimal grammatical adjustments when strictly necessary.
 
\emph{Contradiction Validation.} Each original--mutated claim pair is evaluated using a pretrained NLI model \cite{he_debertav3_2023}. We compute contradiction scores in both directions and retain only claims whose average contradiction score exceeds a threshold of 0.9. This filters out cases where antonym substitution fails to invert meaning cleanly or produces ambiguous statements.
 
\paragraph{In-Lexicon Argument Substitution.} Many claims relate to specific data arguments---attributes, categorical values, or numerical quantities. We identify candidate arguments by generating a structured summary of each visualization's underlying dataset, including attribute names, categorical values, and numerical ranges. An LLM tags spans in each claim corresponding to attributes, values, and numbers using XML-style markers encoding the associated attribute name. This tagging procedure is shared across both in-lexicon and out-of-lexicon substitution strategies.
 
In-lexicon substitution replaces one or more tagged arguments with an alternative from the same dataset lexicon. We apply two forms: categorical value substitution, in which a categorical value is replaced with a different value from the same attribute; and numerical substitution, in which a numerical value is replaced with a different value within the observed range of the attribute. We do not substitute attribute names, as doing so can unintentionally produce true claims when attributes share similar distributions. Only mutated claims exhibiting high NLI contradiction scores ($> 0.9$) are retained and labeled \textsc{False}.
 
\paragraph{Out-of-Lexicon Argument Substitution.} This strategy generates \textsc{NEI} claims by replacing tagged arguments with values or attributes absent from the underlying dataset, introducing references that cannot be verified or refuted. We apply two types: attribute substitution, replacing an attribute name with one not observed in the dataset; and categorical value substitution, replacing a value with one not observed. We do not substitute numerical values, as this can only produce false claims.
 
NLI-based validation ensures that the resulting claim is neither supported nor contradicted by the original source text. We compute entailment and contradiction scores and retain only claims where both fall below a threshold of 0.2, removing cases where the substitution introduces an unintended true or false statement.
 
\subsection{Human Validation Details}
 
A total of 13 annotators with expertise in data visualization participated in the validation study, producing 475 annotations across a sample of 294 claims, with up to three independent annotations per claim. Annotators performed the same task defined in \S\ref{sec:task}: interacting with the visualizations as needed and assigning one of the three labels based solely on the available visual evidence. They were primed with example tasks and given feedback before starting annotation on real samples. Annotators were blind to the original dataset labels and to the construction process.
 
Raw agreement between human annotators and the original labels is 81.3\%. We additionally applied a Bayesian Dawid--Skene model \cite{dawid_maximum_1979, paun_comparing_2018} to the annotation data, inferring latent labels solely from annotator responses. The inferred class prevalence closely matches the dataset distribution: estimated proportions for \textsc{True} and \textsc{False} claims differ by less than 3 percentage points from the ground truth, while NEI prevalence is recovered within the model's uncertainty bounds. The posterior mean accuracy of inferred labels relative to the dataset labels is 66.2\% (95\% credible interval: [59.5\%, 71.8\%]), with an MAP estimate of 77.6\%.

\section{Dataset Construction Method Example}

This appendix illustrates the complete pipeline for constructing claims from human-authored analytical narratives, using a real example from the corpus.

\begin{figure*}
\begin{center}
  \includegraphics[width=\textwidth]{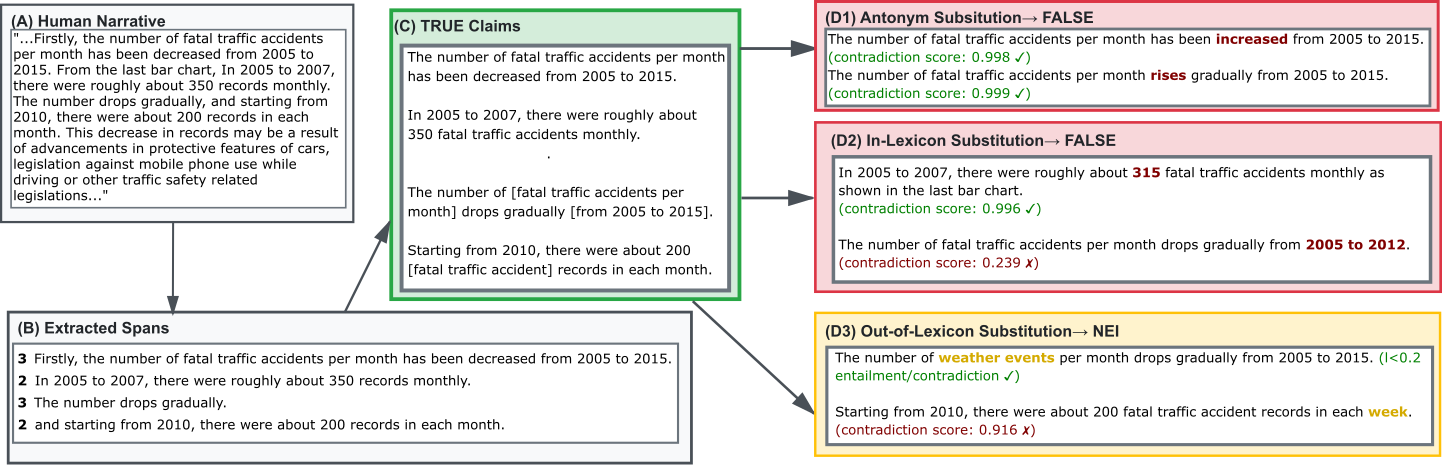}
  \label{fig:method}
  \caption{Worked example of the dataset construction pipeline. Starting from human-authored analytical text about traffic accidents (A), we extract semantically labeled spans (B), decompose them into atomic propositions with resolved references (C), and generate False claims through antonym substitution (D1) and in-lexicon argument replacement (D2), as well as NEI claims through out-of-lexicon substitution (D3). NLI scores validate that mutations produce the intended semantic relationships.}
\end{center}
\end{figure*}

\subsection{Original Text}

\textit{"...Firstly, the number of fatal traffic accidents per month has been decreased from 2005 to 2015. From the last bar chart, In 2005 to 2007, there were roughly about 350 records monthly. The number drops gradually, and starting from 2010, there were about 200 records in each month. This decrease in records may be a result of advancements in protective features of cars, legislation against mobile phone use while driving or other traffic safety related legislations..."}

\subsection{Span Extraction and Semantic Labeling}

Candidate spans are extracted from the narrative and labeled according to their semantic level (Level 2: statistical/relational, Level 3: perceptual/cognitive):

\vspace{0.3cm}
\noindent\textbf{Label 3:} ``Firstly, the number of fatal traffic accidents per month has been decreased from 2005 to 2015.''

\vspace{0.2cm}
\noindent\textbf{Label 2:} ``In 2005 to 2007, there were roughly about 350 records monthly.''

\vspace{0.2cm}
\noindent\textbf{Label 3:} ``The number drops gradually.''

\vspace{0.2cm}
\noindent\textbf{Label 2:} ``and starting from 2010, there were about 200 records in each month.''

\subsection{Decomposition and Decontextualization}

Each span is decomposed into atomic, verifiable propositions with all implicit references resolved:

\vspace{0.3cm}
\noindent\textbf{Span:} ``Firstly, the number of fatal traffic accidents per month has been decreased from 2005 to 2015.''

\noindent\texttt{Proposition: ``The number of fatal traffic accidents per month has been decreased from 2005 to 2015.''}

\vspace{0.3cm}
\noindent\textbf{Span:} ``In 2005 to 2007, there were roughly about 350 records monthly.''

\noindent\texttt{Proposition: ``In 2005 to 2007, there were roughly about 350 fatal traffic accidents monthly [as shown in the last bar chart].''}

\vspace{0.3cm}
\noindent\textbf{Span:} ``The number drops gradually.''

\noindent\texttt{Proposition: ``The number of [fatal traffic accidents per month] drops gradually [from 2005 to 2015].''}

\vspace{0.3cm}
\noindent\textbf{Span:} ``and starting from 2010, there were about 200 records in each month.''

\noindent\texttt{Proposition: ``Starting from 2010, there were about 200 [fatal traffic accident] records in each month.''}

\subsection{Claim Mutation: Contrary Swap}

Antonym substitution generates false claims by reversing directional language. NLI contradiction scores validate semantic opposition:

\vspace{0.3cm}
\noindent\textbf{Input:} The number of fatal traffic accidents per month has been \textbf{decreased} from 2005 to 2015.

\noindent\textbf{Mutated (False):} The number of fatal traffic accidents per month has been \textbf{increased} from 2005 to 2015. \textit{(contradiction score: 0.998)}

\vspace{0.3cm}
\noindent\textbf{Input:} The number of fatal traffic accidents per month \textbf{drops} gradually from 2005 to 2015.

\noindent\textbf{Mutated (False):} The number of fatal traffic accidents per month \textbf{rises} gradually from 2005 to 2015. \textit{(contradiction score: 0.999)}

\subsection{Claim Mutation: In-Lexicon Argument Swap}

Arguments (dates, values) are replaced with alternatives from the dataset to generate false claims:

\vspace{0.3cm}
\noindent\textbf{Mutated (False):} In \textbf{2008 to 2007}, there were roughly about 350 fatal traffic accidents monthly as shown in the last bar chart. \textit{(Contradiction score: 0.984)}

\vspace{0.2cm}
\noindent\textbf{Mutated (False):} In 2005 to 2007, there were roughly about \textbf{315} fatal traffic accidents monthly as shown in the last bar chart. \textit{(Contradiction score: 0.996)}

\vspace{0.2cm}
\noindent\textbf{Mutated (False):} Starting from 2010, there were about \textbf{50} fatal traffic accident records in each month. \textit{(Contradiction score: 0.068 - rejected)}

\subsection{Claim Mutation: Out-of-Lexicon Argument Swap}

Arguments not present in the dataset are substituted to generate NEI (Not Enough Information) claims:

\vspace{0.3cm}
\noindent\textbf{Mutated (NEI):} The number of \textbf{road incidents} per month drops gradually from 2005 to 2015. \textit{(Entailment score: 0.885 - rejected)}

\vspace{0.2cm}
\noindent\textbf{Mutated (NEI):} Starting from 2010, there were about 200 fatal traffic accident records in each \textbf{week}. \textit{(Entailment score: 0.010, Contradiction score: 0.916 - rejected)}

\vspace{0.3cm}
\noindent\textit{Note: Bold text indicates substituted elements. NLI scores validate that mutations produce the intended semantic relationship (high contradiction for false claims, low entailment/contradiction for NEI claims).}

\section{Dataset Statistics} \label{a:dataset}
 
\begin{figure*}
\begin{center}
  \includegraphics[width=\textwidth]{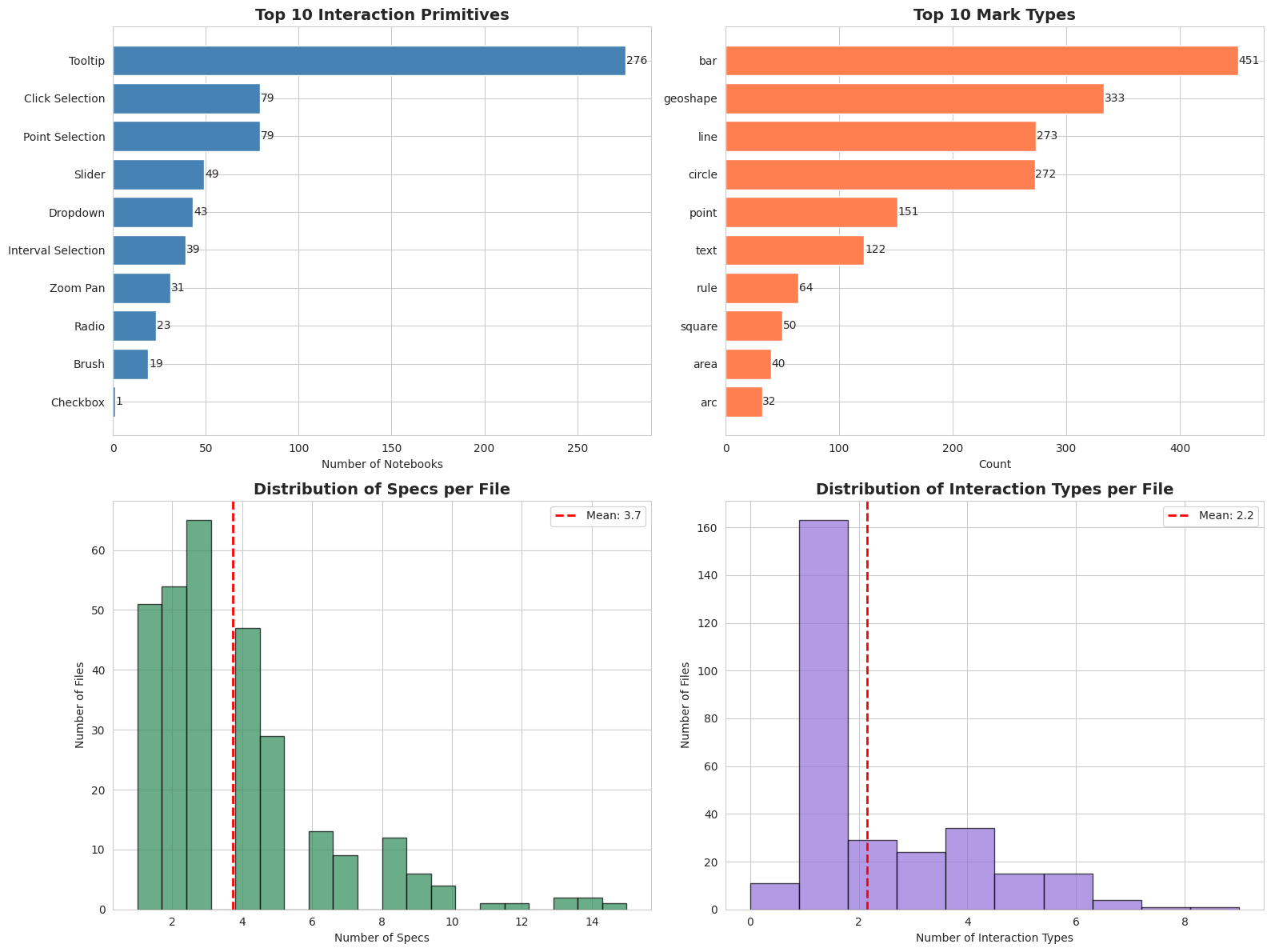}
  \caption{Dataset composition. Top left: distribution of interaction primitives across notebooks. Top right: distribution of Vega-Lite mark types. Bottom left: number of visualization specifications per notebook. Bottom right: number of distinct interaction types per notebook.}
  \label{fig:corpus_stats}
\end{center}
\end{figure*}
 
This appendix provides additional distributional detail complementing the dataset analysis in \S\ref{sec:construction}.
 
\paragraph{Mark types.} The mark-type distribution (Figure~\ref{fig:corpus_stats}, top right) illustrates that InSight is not dominated by a single visualization form. While bars are the most common mark, a substantial fraction of notebooks include lines, geoshapes, circles, text marks, and layered combinations thereof. This diversity is a natural consequence of using Vega-Lite specifications authored by analysts rather than enforcing a fixed chart taxonomy. Importantly, many claims require reasoning across multiple mark types within the same environment (e.g., relating a geographic map to a linked bar chart).
 
\paragraph{Specification and interaction density.} The histograms of visualization specifications per file and interaction types per file (Figure~\ref{fig:corpus_stats}, bottom row) show that most notebooks contain multiple specifications and multiple distinct interaction mechanisms. The long right tails indicate the presence of highly complex environments with many views and interaction affordances. This reinforces that InSight environments are not single-chart snapshots but multi-view analytical spaces in which part of the verification task is identifying where relevant evidence resides.

\clearpage
\onecolumn
\section{Model Evaluation Examples} \label{a:examples}

\begin{figure}[h]
\begin{center}
  \includegraphics[width=\textwidth]{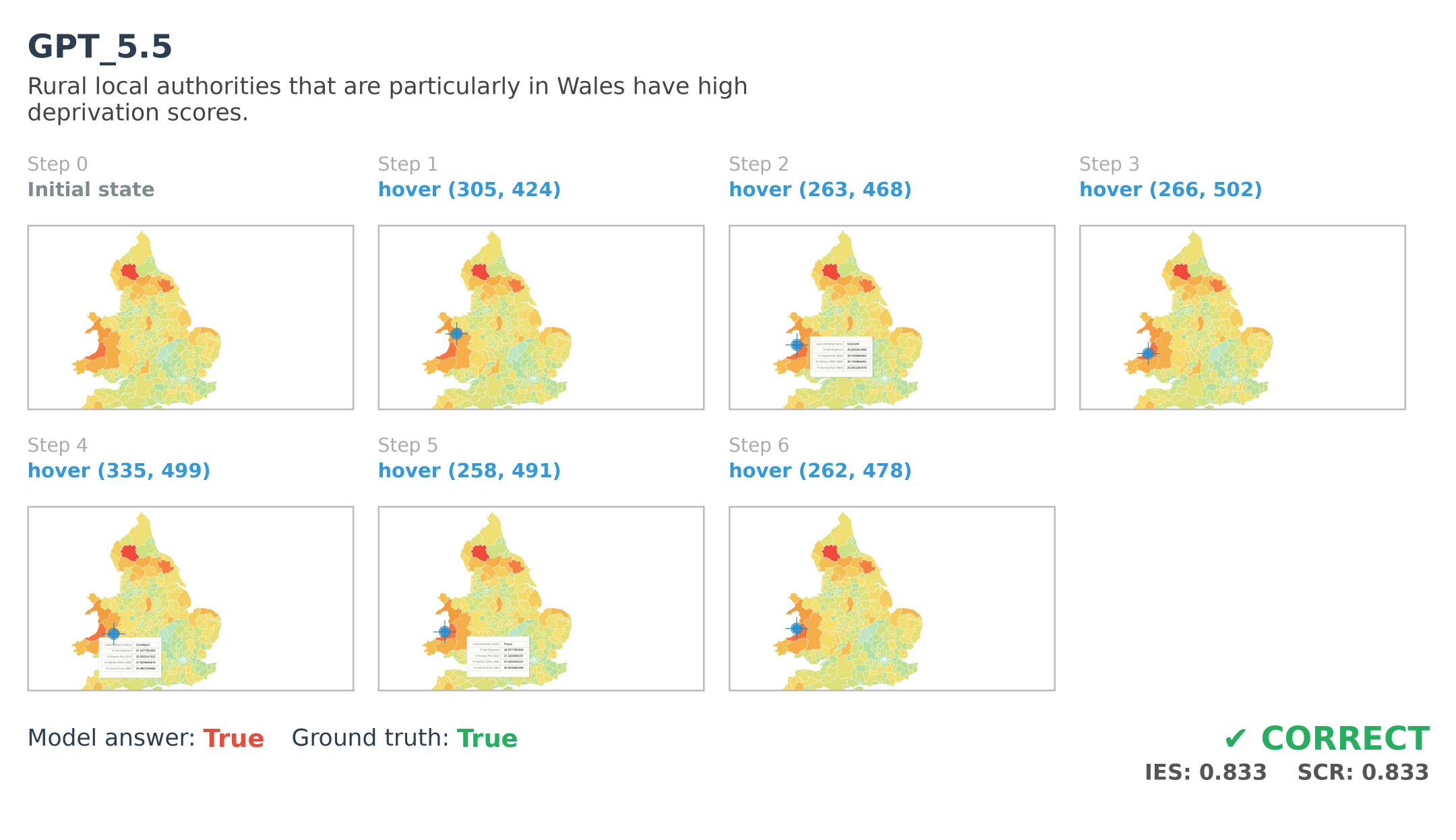}
  \caption{Interaction trace for GPT 5.5 correctly verifying a true claim}
  \label{fig:trace_example}
\end{center}
\end{figure}

\begin{figure}[h]
\begin{center}
  \includegraphics[width=\textwidth]{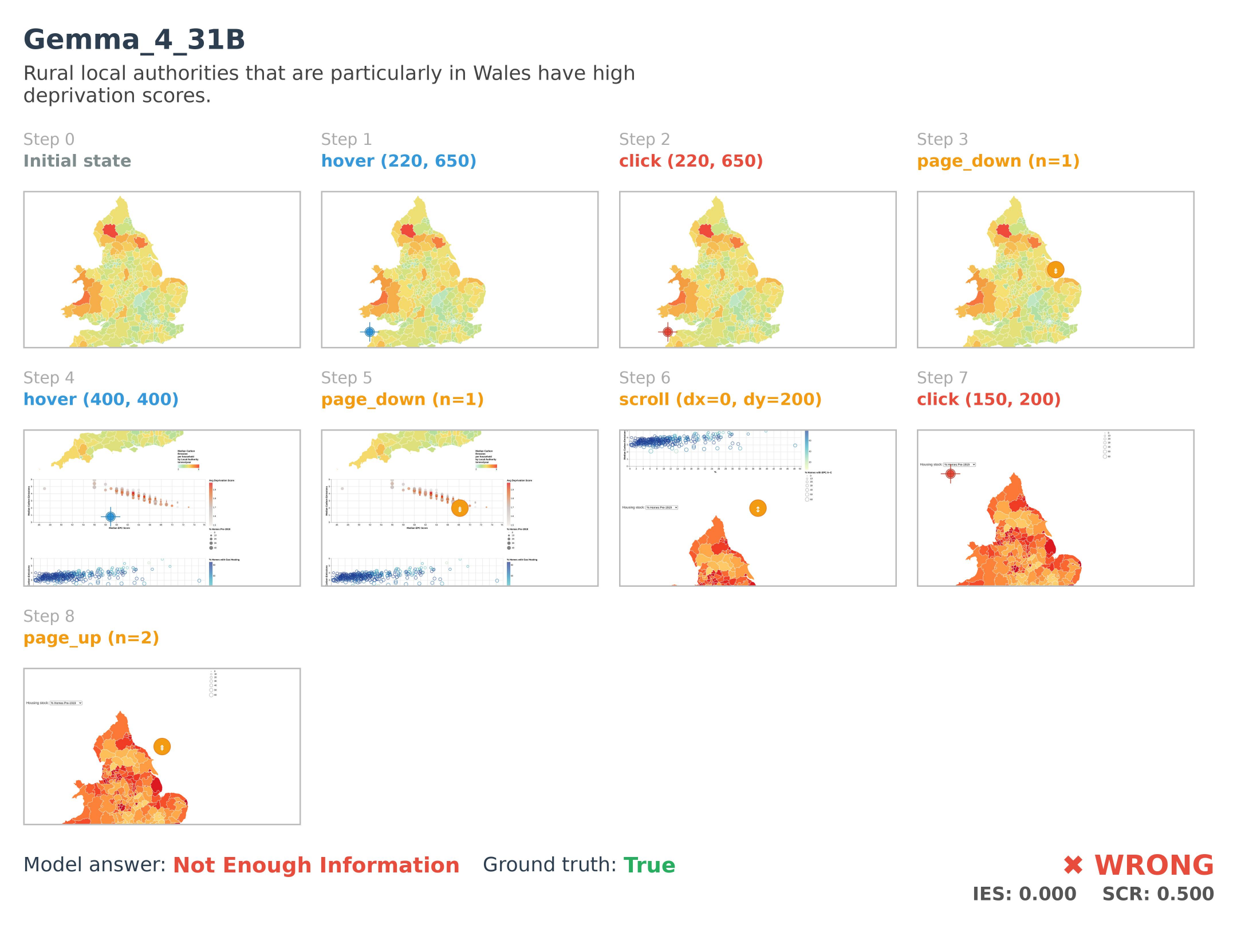}
  \caption{Interaction trace for Gemma 4 31B Flash verifying a true claim, incorrectly labelling the claim as NEI.}
  \label{fig:trace_example}
\end{center}
\end{figure}

\begin{figure}[h]
\begin{center}
  \includegraphics[width=\textwidth]{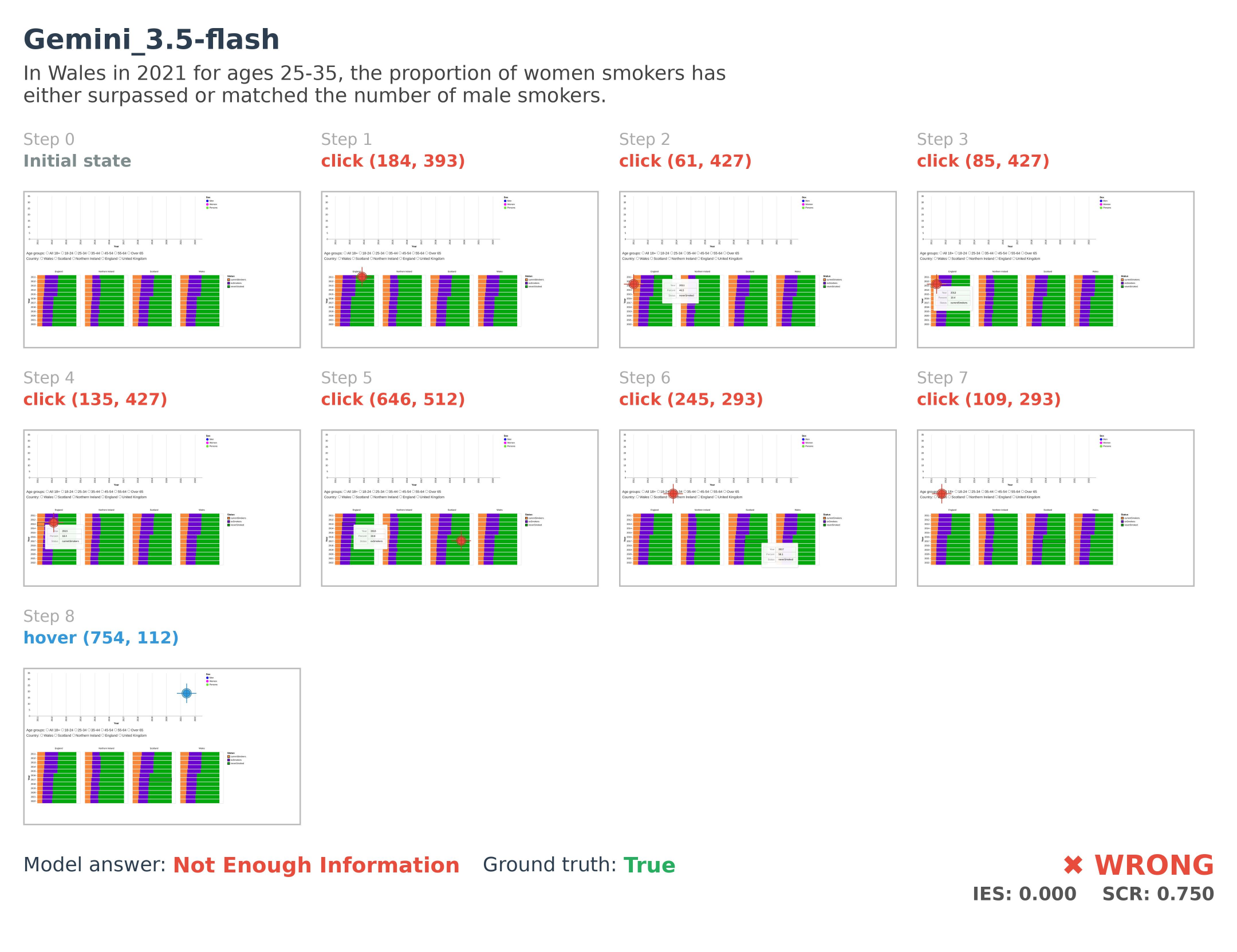}
  \caption{Interaction trace for Gemini 3.5 Flash verifying a true claim, incorrectly labelling the claim as NEI.}
  \label{fig:trace_example}
\end{center}
\end{figure}

\begin{figure}[h]
\begin{center}
  \includegraphics[width=\textwidth]{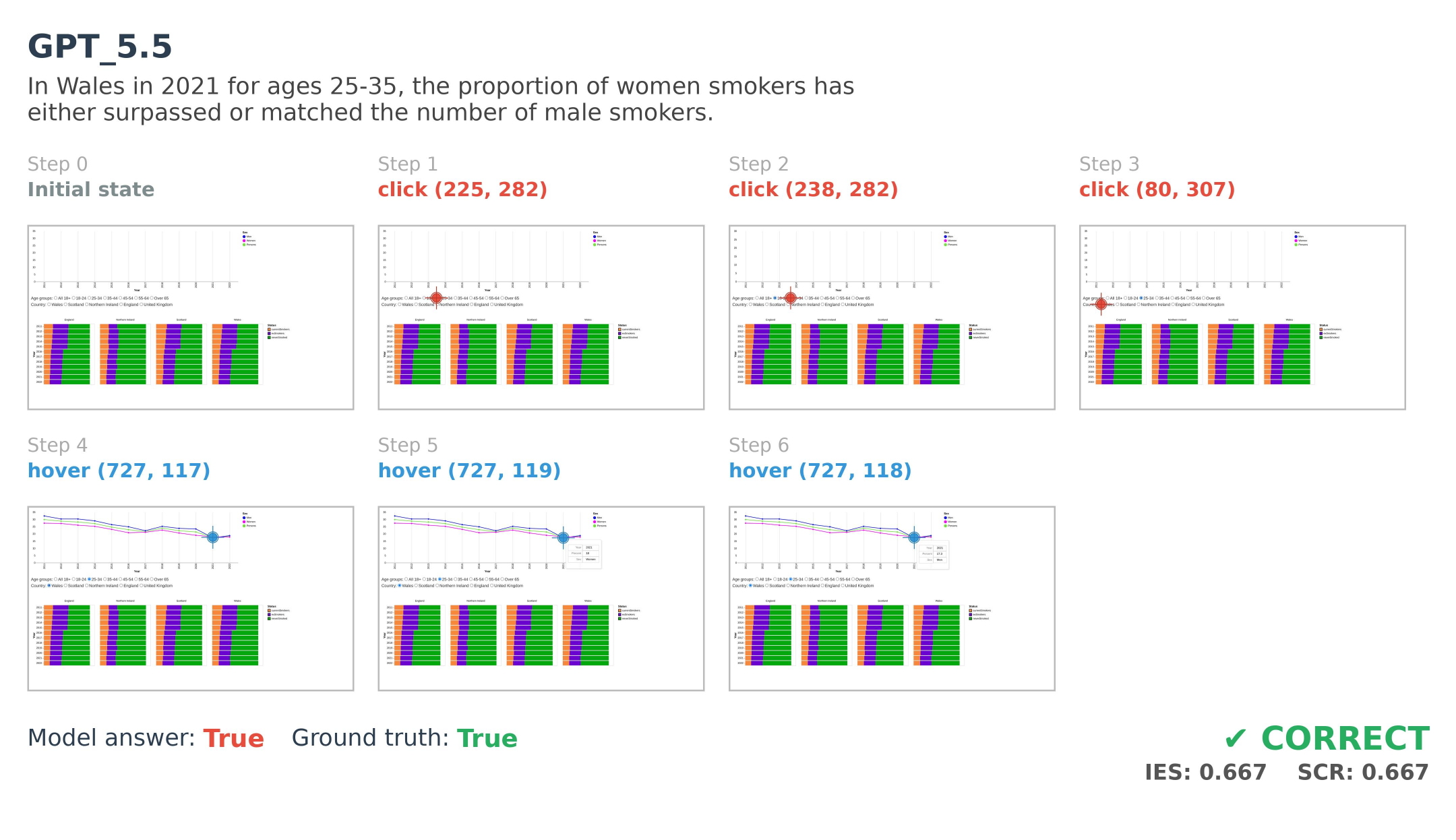}
  \caption{Interaction trace for GPT 5.5 correctly verifying a true claim.}
  \label{fig:trace_example}
\end{center}
\end{figure}

\begin{figure}[h]
\begin{center}
  \includegraphics[width=\textwidth]{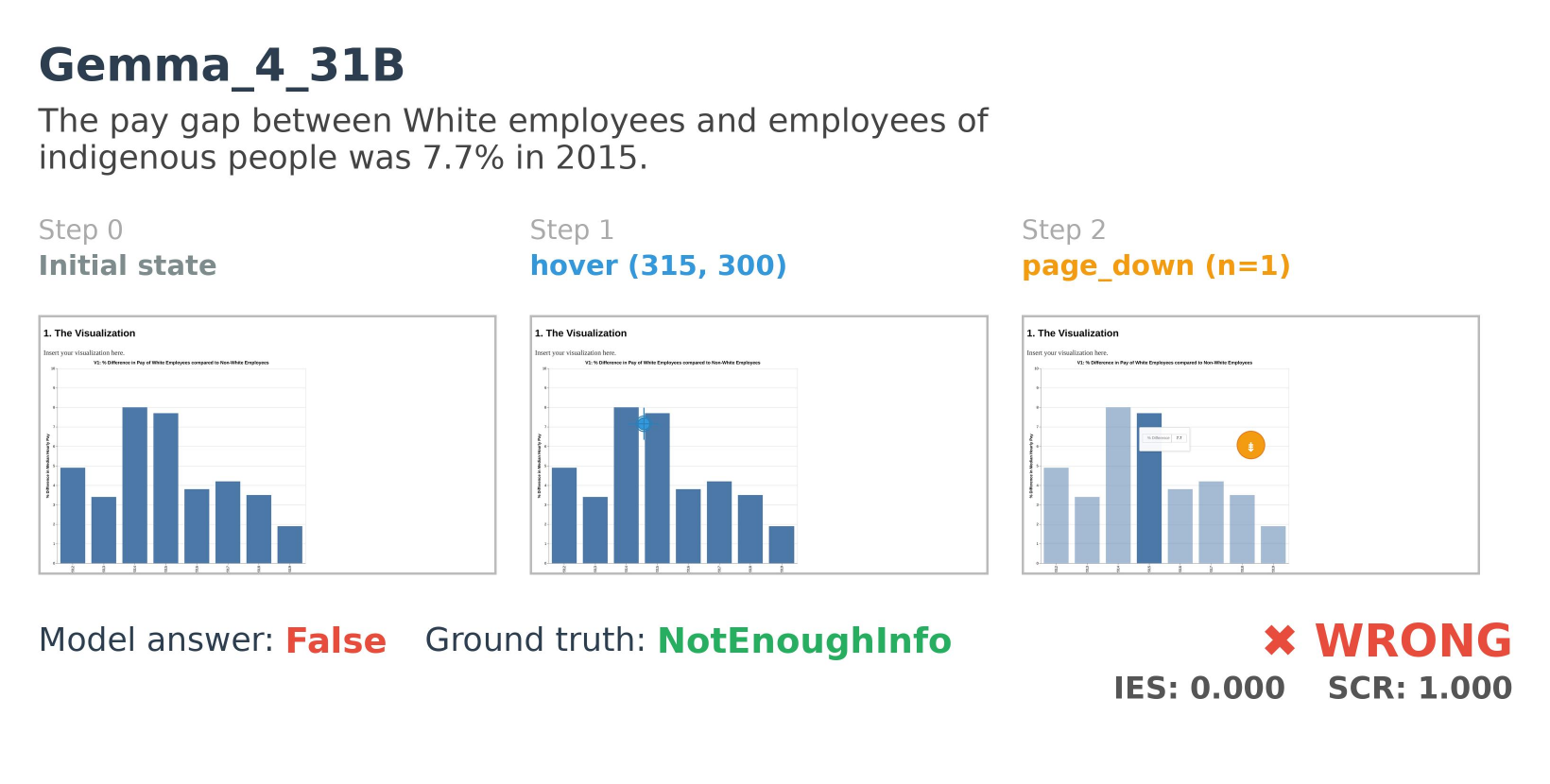}
  \caption{Interaction trace for Gemma 4 31B verifying an NEI claim, incorrectly labelling the claim as False.}
  \label{fig:trace_example}
\end{center}
\end{figure}

\begin{figure}[h]
\begin{center}
  \includegraphics[width=\textwidth]{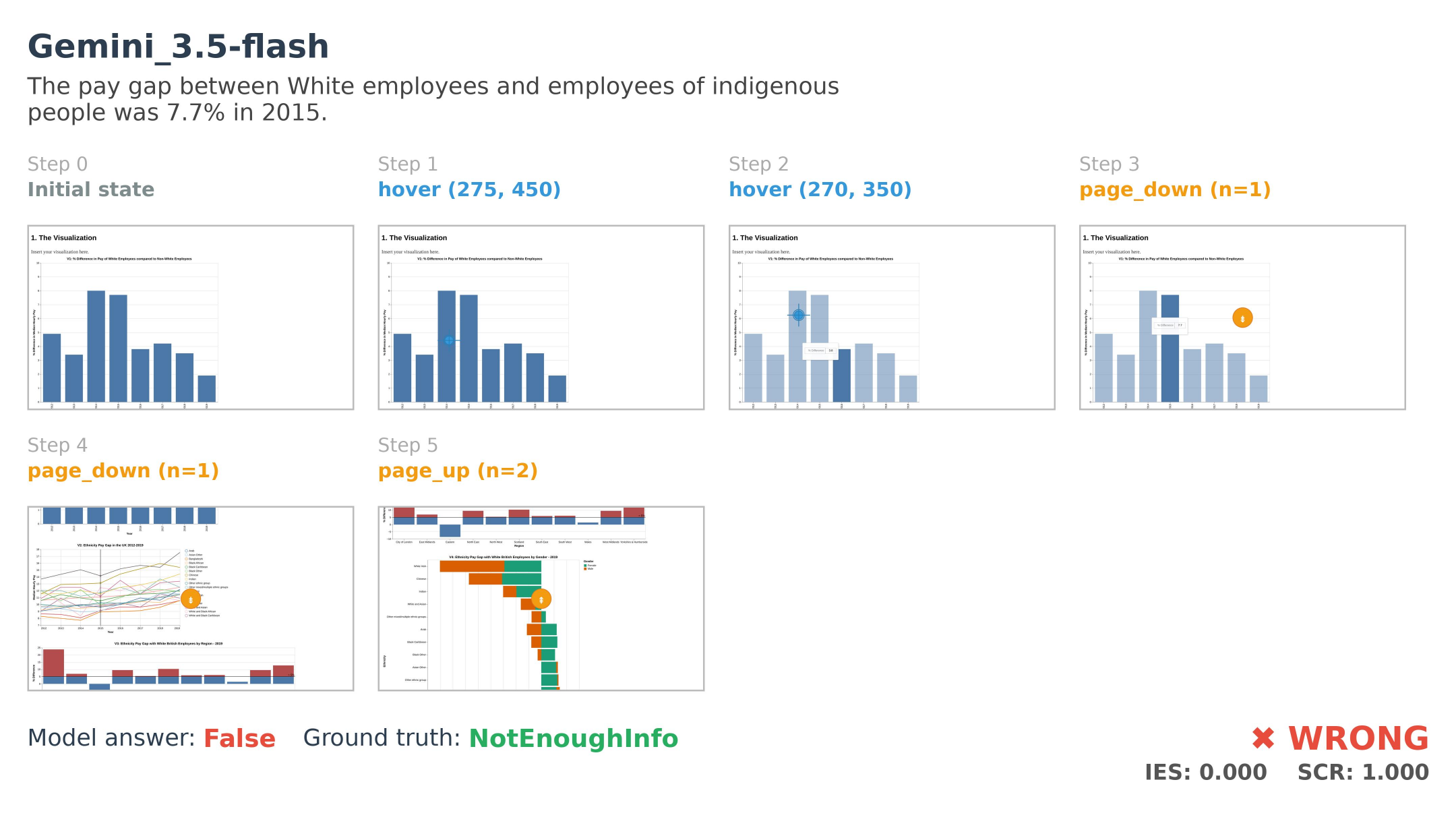}
  \caption{Interaction trace for Gemini 3.5 Flash verifying an NEI claim, incorrectly labelling the claim as False.}
  \label{fig:trace_example}
\end{center}
\end{figure}

\end{document}